# Heart Attack Classification System using Neural Network Trained with Particle Swarm Optimization

Askandar H. Amin[1][0000-0002-1466-4495], Botan K. Ahmed[2][0000-0002-8901-434X], Bestan B. Maaroof[3][0000-0002-6953-8269] and Tarik A. Rashid[4][0000-0002-8661-258X]

[1] Technical College of Informatics, Sulaimani Polytechnic University, Sulaimani, KRG, Iraq
askandar.hamid@spu.edu.iq
[2] Administration and Economics, University of Sulaimani , Sulaimani , KRG, Iraq
botan.ahmed@univsul.edu.iq
[3] IT Department, College of Commerce, University of Sulaimani, Sulaimani , KRG, Iraq
[3] School of Information Technology, Fanshawe College, London , Canada
bestan.maaroof@univsul.edu.iq
[4] Computer Science and Engineering Dept., University of Kurdistan Hewler, Erbil, KRG, Iraq
tarik.ahmed@ukh.edu.krd

**Abstract.** The prior detection of a heart attack could lead to the saving of one's life. Putting specific criteria into a system that provides an early warning of an imminent attack will be advantageous to a better prevention plan for an upcoming heart attack. Some studies have been conducted for this purpose, but yet the goal has not been reached to prevent a patient from getting such a disease. In this paper, Neural Network trained with Particle Swarm Optimization (PSONN) is used to analyze the input criteria and enhance heart attack anticipation. A real and novel dataset that has been recorded on the disease is used. After preprocessing the data, the features are fed into the system. As a result, the outcomes from PSONN have been evaluated against those from other algorithms. Decision Tree, Random Forest, Neural network trained with Backpropagation (BPNN), and Naive Bayes were among those employed. Then the results of 100%, 99.2424%, 99.2323%, 81.3131%, and 66.4141% are produced concerning the mentioned algorithms, which show that PSONN has recorded the highest accuracy rate among all other tested algorithms.

## 1 Introduction

Each year, around twelve million people die due to heart attacks around the globe. The disease affects poor and rich people, both men and women. There are great means specialists set to follow, and the good news is that the disease can be prevented through some steps to help [1]. The main purpose of conducting this research is to showcase where people suffer heart attacks based on some criteria using a recent dataset collected from actual patients who either had a heart attack or not [2].

To contribute, the way this study covers is to enter into the field of Artificial Intelligence for extra accuracy and to save time in assessing such a case in the future application of the model generated by this case study. Furthermore, the result has a higher level of accuracy since the dataset collected from actual cases of the mentioned disease, is based on some features that all people have in common that have a major impact on this type of disease; heart attack. This is a long-lasting disease, it has a long history, and so much research has been done to identify and understand it. The American College of Cardiology (ACC) states that the earliest example of coronary artery disease was discovered in an Egyptian princess, who died of a heart attack near the end of her illness in her forties living between the years 1580 and 1550 B.C. [3].

No final therapy or direct prevention has indeed been released to date to stop heart attacks, and this is an urgent issue when the person affected should be treated at the very early stage of it as quickly as possible. Consequently, more investigation needs to be taken into account to better understand the reasons why one gets a heart attack and to find a quick and effective solution to it for a specific patient, either to monitor the body's activities during old age or in hospitals using artificial intelligence microsystems placed on the patient's body for any of our elderly people. To cope with the disease and/or to specify if one is about to get a heart attack, it is necessary to update the datasets time by time that is used to let the systems that are working concerning that disease be aware of the changes happening in the human body and to treat the algorithms more accurately to prevent someone from dying.

The system testing the dataset that has been used is very user-friendly and has an upgraded hybrid algorithm, combining two great algorithms into one, better to work on datasets for a realistic outcome. So in the upcoming sections, it will be discussed how the algorithms, which have been combined into one piece, are used in a prepared system to train and test such new or any datasets to collect results. In our case, is the patient close to getting a heart attack or not. After that, the methodology for using the system and the grabbed results will be presented in a discussion.

The rest of the paper is divided into the following sections: Literature Review is explained in Section 2, Methodology is described in Section 3. Section 4, displays the results, and finally, the main points are concluded in Section 5.

## 2    Literature Review

As mentioned earlier in this paper, heart attack early detection has continuously gained the attention of researchers, especially those in the artificial intelligence field. Various studies and research have been conducted for this purpose, as early detection of a heart attack is a huge step toward saving many people's lives. Research conducted by Feshki and Shijani [4] presented the Particle Swarm Optimization system (PSO) and feed-forward neural network for feature ranking on affected factors. The proposed system had the impact of the reduction of thirteen effective cardiovascular factors to eight optimized features. As a result of the assessment of the selected features, the system showed a high accuracy rate of 91.94%.

Shanthi proposed an adaptive neuro-fuzzy inference system [5]. The proposed system relied on some features based on the lifestyle of the patient taken as input to the system and then neuro-fuzzy inference predicts the probability of cardiovascular disease. Ali et al. [6] proposed an automated diagnostic system based on deep neural networks (DNN) for making predictions for heart disease. Then $X^2$ statistical model was conducted for feature refinement. The Cleveland dataset was used for testing and training. The results claimed better performance compared to the previously presented neural networks.

Very recent work has been done by Wang et al. [7] to predict heart attack in stroke patients. They used an undersampling, clustering-oversampling algorithm (UCO) for imbalanced data processing as well as feature extraction. The output from UCO120 was fed into five different classifiers, and the test results showed that the random forest had gained an accuracy of 70.29%, which was the best performance among the other classifiers. Another study conducted by Ali et al. [8] used two support vector machines (SVM). The first linear SVM was used for feature selection as well as the removal of extraneous features. The second SVM was used for prediction.

Javeed et al.[9] proposed an intelligent hybridized random search algorithm (RSA) with random forest (RF) as an intelligent learning system for heart failure detection. In the proposed RSA-RF system, RSA was used for feature selection and RF was used for prediction based on the selected features from RSA. Another recent piece of research work has been done by Maghdid et al. [2]. The same dataset was used as we used in this current work. They used three techniques; BPNN was one of them, but in the current work, a PSONN [10] is proposed as well. They used two more techniques, the Fuzzy Inference System (FIS) and Adaptive Neuro-Fuzzy System (ANFIS) as predictors for heart attack. The experimental results showed ANFIS to be the most efficient model compared with the other two techniques.

## 3  Methodology

For this research, the dataset collected by Maghdid et al. [2] is used to detect heart attack systems which are suggested based on the particle swarm optimization neural network (PSONN). This process falls into two steps; firstly, the neural network with two hidden layers is structured to receive the data set [2]. The neural network's weights and biases will be optimized using particle swarm optimization (PSO). Secondly, the PSONN is tested with the dataset to see how well it has been trained [10].

### 3.1 Dataset

The dataset used for this paper is the same one that was collected from the Surgical Specialty Cardiac Centre's "Directorate of Health-Erbil" Kurdistan Regional Government Iraq, 2018. The size of the provided dataset is 1319 with 8 attributes as shown in Table 1; age, gender, pulse, blood pressure systolic, blood pressure diastolic, glucose, CK-MB, and troponin with only one output; either indicates the existence of a heart attack (positive) or not (negative). The actual data set can be obtained here [11].

**Table 1.** Data Attributes & Their Descriptions

| Attributes | Descriptions |
| --- | --- |
| Age | Age of the patient in the year |
| Gender | Gender of the patient (male or female) |
| Heart rate | Maximum heart rate achieved 10 to 160-180 |
| Systolic blood pressure | Resting systolic blood pressure (in mm Hg on ad mission to the hospital) (70-190) |
| Diastolic blood pressure | Resting diastolic blood pressure (in mm Hg on ad mission to the hospital) (40-100) |
| Blood sugar | (Blood sugar > 120 mg/dl) (1-900mg/dl) |
| CK-MB | Enzyme CK-MB (male upto-6.22 female upto-4.88) It is Creatine Kinase Enzymes |
| Troponin | Enzyme Troponin (0.0-0.014) |

According to the provided information, the medical dataset is to classify whether it is a heart attack or not. The data has been encoded in an excel sheet as follows: the male is set to 1 and the female to 0. The glucose column is set to 1 if it is > 120, otherwise 0. As for the output, positive is set to 1 and negative to 0 (see Figure 1).

| | A | B | C | D | E | F | G | H | I | J |
| --- | --- | --- | --- | --- | --- | --- | --- | --- | --- | --- |
| 1 | Age | Gender | Pulse | Pressure high | Pressure low | Glucose | CK-MB | Troponin | Target | Target |
| 2 | 64 | 1 | 66 | 160 | 83 | 1 | 1.8 | 0.012 | 0 | negative |
| 3 | 21 | 1 | 94 | 98 | 46 | 1 | 6.75 | 1.06 | 1 | positive |
| 4 | 55 | 1 | 64 | 160 | 77 | 1 | 1.99 | 0.003 | 0 | negative |
| 5 | 64 | 1 | 70 | 120 | 55 | 1 | 13.87 | 0.122 | 1 | positive |
| 6 | 55 | 1 | 64 | 112 | 65 | 1 | 1.08 | 0.003 | 0 | negative |
| 7 | 58 | 0 | 61 | 112 | 58 | 0 | 1.83 | 0.004 | 0 | negative |
| 8 | 32 | 0 | 40 | 179 | 68 | 0 | 0.71 | 0.003 | 0 | negative |
| 9 | 63 | 1 | 60 | 214 | 82 | 0 | 300 | 2.37 | 1 | positive |
| 10 | 44 | 0 | 60 | 154 | 81 | 1 | 2.35 | 0.004 | 0 | negative |
| 11 | 67 | 1 | 61 | 160 | 95 | 0 | 2.84 | 0.011 | 0 | negative |
| 12 | 44 | 0 | 60 | 166 | 90 | 0 | 2.39 | 0.006 | 0 | negative |
| 13 | 63 | 0 | 60 | 150 | 83 | 1 | 2.39 | 0.013 | 0 | negative |
| 14 | 64 | 1 | 60 | 199 | 99 | 0 | 3.43 | 5.37 | 1 | positive |
| 15 | 54 | 0 | 94 | 122 | 67 | 0 | 1.42 | 0.012 | 0 | negative |
| 16 | 47 | 1 | 76 | 120 | 70 | 1 | 2.57 | 0.003 | 0 | negative |
| 17 | 61 | 1 | 81 | 118 | 66 | 1 | 1.49 | 0.017 | 1 | positive |
| 18 | 86 | 0 | 73 | 114 | 68 | 0 | 1.11 | 0.776 | 1 | positive |
| 19 | 45 | 0 | 70 | 100 | 68 | 0 | 0.606 | 0.004 | 0 | negative |
| 20 | 37 | 0 | 72 | 107 | 86 | 1 | 2.89 | 0.003 | 0 | negative |

**Fig. 1.** A snapshot of the Medical Dataset in Excel (20 out of 1319)

### 3.2 Neural Network Trained with Particle Swarm Optimization

An elaborated and inherited network based on particle swarm optimization and neural network (PSONN). The proposed PSONN was developed from a neural network (NN) model trained by particle swarm optimization (PSO) for optimizing the weights and biases of the neural network. PSO is used to obtain the optimum weights, and then the optimized weights are fed into the neural network for decision making in the testing phase. Then, afterward, these weights are fed into the NN to test the system (PSONN). The flow chart for PSONN algorithms is shown below (Figure 2).

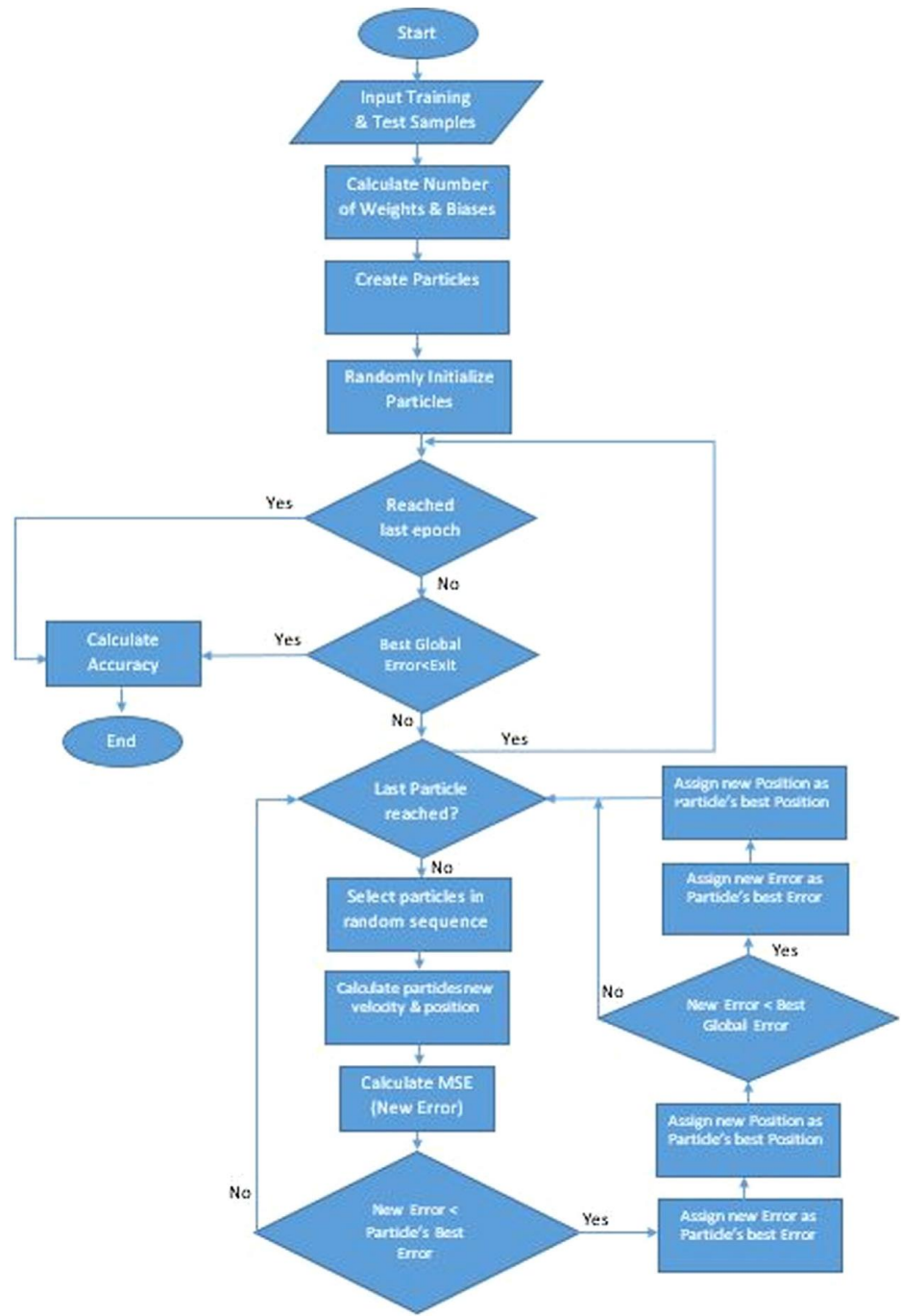

**Fig. 2.** Flow chart for PSONN algorithm [10].

## 4  Results and Discussions

The results obtained by the algorithm are as follows for the PSONN model described in the methodology section. Table 2 displays the results of the PSONN model having different epochs applied.

**Table 2.** Results of the PSONN model for different epochs.

| ID | # of epochs | # of features | # of observations | # of hidden layers | MAE | RMSE | Training accuracy |
|---|---|---|---|---|---|---|---|
| | | | Train phase | | | | |
| 1 | 200 | 8 | 1055 | 2 | 0 | 0 | 100.00% |
| 2 | 500 | 8 | 1055 | 2 | 0 | 0 | 100.00% |
| 3 | 700 | 8 | 1055 | 2 | 0 | 0 | 100.00% |
| | | | Test phase | | | | |
| ID | # of epochs | # of features | # of observations | # of hidden layers | MAE | RMSE | Testing accuracy |
| 1 | 200 | 8 | 264 | 2 | 0 | 0 | 100.00% |
| 2 | 500 | 8 | 264 | 2 | 0 | 0 | 100.00% |
| 3 | 700 | 8 | 264 | 2 | 0 | 0 | 100.00% |

Table 2 indicates the accuracy of PSONN with having different epochs it achieved 100% accuracy during the training and testing phases.

| Confusion Matrix | | | | | | | |
|---|---|---|---|---|---|---|---|
| PSONN – 700 Epochs | | | | | | | |
| Training phase | | | | Testing phase | | | |
| | | True labels | | | | True labels | |
| | | 1 | 2 | | | 1 | 2 |
| Predicted labels | 1 | 398 | 0 | Predicted labels | 1 | 111 | 0 |
| | 2 | 0 | 657 | | 2 | 0 | 153 |

**Fig. 3.** Confusion matrix for 700 epochs for PSONN model.

Figure 3 demonstrates the accuracy of PSONN during the training and testing phases. The above outcomes were collected when PSONN was configured to run for 700 epochs. The result is that increasing the number of epochs will significantly improve the performance of the PSONN. However, it increases the time and computational power when generating the model. PSONN was tested when there were 700 epochs. It was capable of correctly classifying all the observations during the training and testing phases without any errors. Even after 10,000 epochs of the heart attack dataset, the PSONN algorithm achieves 100% accuracy. Consequently, some other techniques are applied to the same data to achieve a proper result that satisfies researchers.

The Weka tool has been used as a machine learning tool and techniques have been developed using some approaches on how to best fit the data in the tool [12-14]. After loading the data into the tool, it is normalized for better comparison between the features of the dataset. Then four techniques are applied by dividing the dataset into 70% for training and 30% for testing. The results vary; first, the Decision Tree algorithm includes two hidden layers, each having five nodes, and for 500 epochs, the accuracy is the highest compared to the other three algorithms, which average 99.2424%. Second, with slightly lower accuracy, is the Random Forest technique, with 99.2323%. Third, the Backpropagation Neural Network received 81.3131% of the accuracy, and Naive Bayes came in last with 66.4141% of the accuracy. All the results are shown in the below figures along with their confusion matrixes included in them.

```
=== Summary ===

Correctly Classified Instances         393               99.2424 %
Incorrectly Classified Instances         3                0.7576 %
Kappa statistic                          0.9839
Mean absolute error                      0.0129
Root mean squared error                  0.0871
Relative absolute error                  2.7335 %
Root relative squared error             17.9625 %
Total Number of Instances              396

=== Detailed Accuracy By Class ===

                 TP Rate  FP Rate  Precision  Recall  F-Measure  MCC    ROC Area  PRC Area  Class
                 0.988    0.000    1.000      0.988   0.994      0.984  0.990     0.995     positive
                 1.000    0.012    0.980      1.000   0.990      0.984  0.990     0.970     negative
Weighted Avg.    0.992    0.005    0.993      0.992   0.992      0.984  0.990     0.986

=== Confusion Matrix ===

   a   b   <-- classified as
 244   3 |   a = positive
   0 149 |   b = negative
```

**Fig. 4.** Accuracy of Weka Decision Tree on the heart attack dataset.

Figure 4 illustrates that the Decision Tree algorithm gives the best accuracy for the heart attack dataset which is 99.2424% that correctly classified 393 out of 396 in the testing phase; to mention, all the negative cases of 149 cases were correctly categorized.

```
=== Summary ===

Correctly Classified Instances         389               98.2323 %
Incorrectly Classified Instances         7                1.7677 %
Kappa statistic                          0.9626
Mean absolute error                      0.03
Root mean squared error                  0.1051
Relative absolute error                  6.3392 %
Root relative squared error             21.6845 %
Total Number of Instances              396

=== Detailed Accuracy By Class ===

                 TP Rate  FP Rate  Precision  Recall  F-Measure  MCC    ROC Area  PRC Area  Class
                 0.976    0.007    0.996      0.976   0.986      0.963  0.997     0.998     positive
                 0.993    0.024    0.961      0.993   0.977      0.963  0.997     0.988     negative
Weighted Avg.    0.982    0.013    0.983      0.982   0.982      0.963  0.997     0.994

=== Confusion Matrix ===

   a   b   <-- classified as
 241   6 |   a = positive
   1 148 |   b = negative
```

**Fig. 5.** Accuracy of Weka Random Forest on the heart attack dataset.

Figure 5 clarifies that the Random Forest algorithm gives another great accuracy for the heart attack dataset which is 99.2323%. It categorized 148 out of 149 for negative cases and missed only 6 cases out of 247 positive circumstances.

```
=== Summary ===

Correctly Classified Instances         322               81.3131 %
Incorrectly Classified Instances        74               18.6869 %
Kappa statistic                          0.6071
Mean absolute error                      0.2562
Root mean squared error                  0.3544
Relative absolute error                 54.1912 %
Root relative squared error             73.1281 %
Total Number of Instances              396

=== Detailed Accuracy By Class ===

                 TP Rate  FP Rate  Precision  Recall  F-Measure  MCC     ROC Area  PRC Area  Class
                 0.830    0.215    0.865      0.830   0.847      0.608   0.896     0.945     positive
                 0.785    0.170    0.736      0.785   0.760      0.608   0.896     0.785     negative
Weighted Avg.    0.813    0.198    0.816      0.813   0.814      0.608   0.896     0.885

=== Confusion Matrix ===

   a   b   <-- classified as
 205  42 |   a = positive
  32 117 |   b = negative
```

**Fig. 6.** Accuracy of Weka Backpropagation Neural Network on the heart attack dataset.

Figure 6 proves that the Backpropagation Neural Network algorithm gives the accuracy of 81.3131% for the heart attack dataset with two hidden layers of five nodes each. It incorrectly classified 42 positives and 32 negatives out of 396 cases in total for the testing phase.

```
=== Summary ===

Correctly Classified Instances         263               66.4141 %
Incorrectly Classified Instances       133               33.5859 %
Kappa statistic                          0.3921
Mean absolute error                      0.3343
Root mean squared error                  0.5481
Relative absolute error                 70.7018 %
Root relative squared error            113.0935 %
Total Number of Instances              396

=== Detailed Accuracy By Class ===

                 TP Rate  FP Rate  Precision  Recall  F-Measure  MCC     ROC Area  PRC Area  Class
                 0.462    0.000    1.000      0.462   0.632      0.494   0.840     0.914     positive
                 1.000    0.538    0.528      1.000   0.691      0.494   0.840     0.681     negative
Weighted Avg.    0.664    0.203    0.823      0.664   0.654      0.494   0.840     0.826

=== Confusion Matrix ===

   a   b   <-- classified as
 114 133 |   a = positive
   0 149 |   b = negative
```

**Fig. 7.** Accuracy of Weka Naive Bayes on the heart attack dataset.

Figure 7 shows that the Naive Bayes algorithm gives the lowest accuracy for the heart attack dataset which is 66.4141%. Surprisingly, it covered all the 149 negative cases but missed 133 of the positive cases out of 247 positive circumstances.

## 5 Conclusion

In this paper, PSONN was used to classify heart attacks. A relatively big dataset was used for classification purposes. PSONN outperformed other competitive algorithms, such as Decision Tree, Random Forest, BPNN, and Naive Bayes. Thus, the PSONN model can be relied on by hospitals to predict patients' states according to the experimental results that we obtained. For future work, this algorithm could be tested with different datasets to ensure and verify its performance.